\documentclass[10pt]{article} 
\usepackage[preprint]{tmlr}
\usepackage{graphicx}
\usepackage{natbib}
\usepackage{parskip}
\usepackage{microtype}
\usepackage{subcaption}
\usepackage{booktabs}
\usepackage{amsmath}
\usepackage{amssymb}
\usepackage{mathtools}
\usepackage{amsthm}
\usepackage{hyperref}
\usepackage{url}
\usepackage{booktabs}
\usepackage{amsfonts}
\usepackage{nicefrac}
\usepackage{microtype}
\usepackage{xcolor}
\usepackage{graphicx}
\usepackage{dsfont}
\usepackage{thmtools, thm-restate}
\usepackage{algorithmic}
\usepackage{algorithm}
\usepackage{pythonhighlight}

\usepackage{multirow}
\usepackage{xspace}
\usepackage{enumitem}
\usepackage{fontawesome}

\usepackage{tabularx}
\usepackage{array}
\newcolumntype{C}{>{\centering\arraybackslash}X}

\newcommand{\blockcomment}[1]{}

\newcommand{\T}{\top}

\newcommand{\E}{\mathbb{E}}
\newcommand{\R}{\mathbb{R}}

\newcommand{\cD}{\mathcal{D}}

\renewcommand{\th}{\theta}

\newcommand{\zi}[1]{}

\theoremstyle{plain}

\theoremstyle{definition}

\theoremstyle{remark}

\title{Tracing LLM Behavior to the Training Data with \\Empirical Next-Token Distributions}
\author{\name Zachary Izzo \email zach@nec-labs.com \\
      \addr NEC Labs America}
\date{}

\begin{document}

\maketitle

\begin{abstract}
In this paper, we study the connection between an LLM's output distribution and the data used to train it.
Specifically, we study the degree to which an LLM's next-token distribution agrees with the empirical next-token distribution (ENTD) given the context in the training data.
The ENTD is an appealing target because it is the unrestricted global minimizer of the next-token cross entropy loss used for pretraining, as well as an easily interpretable function of the pretraining corpus.
We find that for a significant fraction of inputs, the LLM's distribution agrees with the ENTD almost perfectly, and the agreement generally increases with model scale and training compute.
Nevertheless, there is a long tail of input sequences where the LLM and ENTD differ significantly, and we examine several possible sources of this discrepancy across the transformer architecture, training procedure, and finite-sample noise in the ENTD estimate itself.
More broadly, we hope our findings will encourage more work on ``data-centric mechanistic interpretability,'' a complement to standard mechanistic interpretability that opens the black box of how model behaviors arise from the data, rather than how they are encoded in the learned weights.
\end{abstract}

\section{Introduction}
The goal of this paper is to enhance our understanding of how LLM capabilities emerge from the training data. It is well-known that the training data are critical for shaping what the model can and cannot do \citep{blevins2022data, xie2023data, xie2023data2, ye2025data}. In spite of this, a more precise, first-principles-based explanation of where model capabilities ``come from'' in the data remains elusive. This paper is an attempt to make progress on this front.

To this end, we study how well a trained transformer approximates the empirical next-token distribution (ENTD) of its data. The ENTD is the global minimizer of the next-token prediction cross-entropy loss used in pretraining, and it is easily interpretable as a fully ``memorized'' model of the data: the predictions for the next token are simply the empirical distribution of next tokens, given the context. The existence of in-context learning capabilities (e.g., induction heads \citep{olsson2022induction}, which can make sensible predictions even on semi-random, completely out-of-distribution inputs) means that the ENTD cannot capture the full capabilities of modern LLMs. Nevertheless, we believe that studying the extent to which an LLM approximates the ENTD is an important initial step to understanding the link between training data and trained model.

We therefore examine how the LLM's predictions relate to the ENTD predictions as training progresses across a wide range of model scales. Our results reveal a complicated picture with deviations between the model and ENTD potentially arising from a variety of sources, including sampling error, optimization bias, optimization variance, and the limitations and inductive biases of the transformer architecture. These results can be viewed as ``scaling laws under a microscope'': the clean, steady decrease in average training loss (which is essentially the mean discrepancy between the model and ENTD) conceals a wide range of per-sample behaviors. Understanding the full breadth of these dynamics is critical for enhancing our understanding of LLMs as a whole.

In summary, we make the following contributions:
\begin{enumerate}
    \item We study how the discrepancy between an LLM's output and the empirical next-token distribution of the training data evolves over the course of training. Unlike classical scaling laws, which are equivalent to the mean discrepancy measured in cross-entropy, we study the \emph{distribution} of discrepancies, revealing a wide range of behaviors from complete convergence to the ENTD to complete deviation.
    \item We examine how different components of the model creation procedure contribute to this discrepancy. These include sampling error, variance in the optimization procedure, and structural properties of the transformer architecture.
    \item We also analyze properties of the data itself which correlate with greater or lesser similarity between the model and the ENTD. In addition to well-established or intuitive effects, we surprisingly find that LLMs struggle to approximate the next-token distributions of frequent examples with a high empirical entropy.
\end{enumerate}

\begin{figure}
    \centering
    \includegraphics[width=0.5\linewidth]{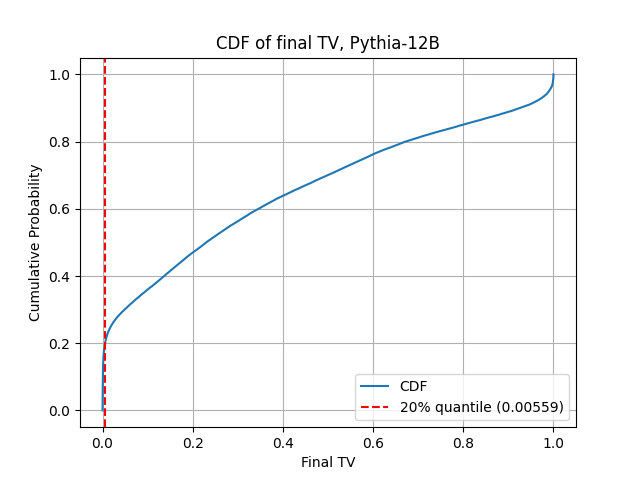}
    \caption{At a very high level, our findings can be summarized by this CDF plot. The interpretation is that the LLM's next-token distribution almost perfectly matches (TV $\leq 0.0056$) the empirical next-token distributions for 20\% of a stratified evaluation sample. The remainder of the paper examines this phenomenon in greater detail, and tries to account for the remaining 80\%.}
    \label{fig:summary}
\end{figure}

\section{Background}
\subsection{Standard ENTD}
We focus on the base model obtained after the pretraining stage. Recent evidence shows that many of the final model's capabilities are learned during pretraining, with common post-training stages improving the sample efficiency of the final model (i.e., the chance that the LLM produces a useful output on the first try, rather than requiring many attempts) rather than fundamentally expanding its reasoning capabilities \citep{yue2025doesrl}.
Thus, for the sake of simplicity in this initial exploration, we restrict our attention to this first important step.

The pretraining phase updates the model weights to minimize the cross-entropy loss on a next-token prediction task. Specifically, let $p_\th(x_{k+1}|x_{1:k})$ be the model's predicted probability that the token $x_{k+1}$ is the next token after the string of $k$ tokens $x_{1:k}$, where the model parameters/weights are collected in $\th$. Pretraining updates $\th$ to approximately minimize
\begin{equation} \label{eq: pretraining loss}
\E_{x_{1:k+1} \sim \cD}[-\log p_\th(x_{k+1} | x_{1:k})],
\end{equation}
where $\cD$ is the training dataset (consisting of a massive number of text documents, represented as strings of tokens) and inputs of variable lengths $x_{1:k+1}$ are sampled from this corpus.
If $p_\th$ can represent arbitrary functions of its input, then we can exactly compute the global minimizer of \eqref{eq: pretraining loss}. It is given by the empirical next-token distribution $p_{\textrm{ENTD}}$ of the training data, which is defined as
\begin{equation} \label{eq: ENTD}
p_{\textrm{ENTD}}(x_{k+1}|x_{1:k}) = \frac{\textrm{count}(x_{1:k+1}, \cD)}{\textrm{count}(x_{1:k}, \cD)},
\end{equation}
i.e., the empirical probability that $x_{k+1}$ follows $x_{1:k}$ in the training data.

\subsection{Training-Aligned ENTD}
During training, the model does not make a prediction on every context present in the training data. Rather, a common procedure (and the one employed by the models we experimented with) is for the data to be loaded in batches consisting of $B$ examples of consecutive tokens of size equal to the model's max context length $C$. If $x^{(b)}_{1:C}$ is one such example, then the loss is computed only on \emph{prefixes} of the example, i.e.,
\[
\sum_{b=1}^B\sum_{i=1}^C -\log p_\th(x^{(b)}_{i+1} | x^{(b)}_{1:i}).
\]
Thus, the minimizer of the loss for the examples which are actually seen and evaluated during training is not the ENTD, but rather an analog (which we dub the \emph{training-aligned empirical next-token distribution} or ENTD-T for short) which respects this batch-aligned structure. Specifically, let $x_{0:N-1}$ denote the full tokenized corpus. Given an input $\hat{x}_{1:k}$ of length $k$ and a next token $\hat{x}_{k+1}$, the ENTD-T is defined as
\[
p_{\textrm{ENTD-T}}(\hat{x}_{k+1}|\hat{x}_{1:k}) = \frac{\#\{0 \leq i \leq N / (C+1) \: | \: x_{(C+1)i: (C+1)i + k + 1} = \hat{x}_{1:k+1}\}}{\#\{0 \leq i \leq N / (C+1) \: | \: x_{(C+1)i: (C+1)i + k} = \hat{x}_{1:k}\}}.
\]
Equivalently, this is the fraction of training context-aligned prefixes starting with $\hat{x}_{1:k}$ where the next token was $\hat{x}_{k+1}$. (Note that training context alignment is a mod-$(C+1)$ quantity since the need for next-token ``targets'' means that each example of length $C$ actually consumes $C+1$ tokens. This is aligned with the training dataloader for the models we tested.) Similarly to how the ENTD was defined only for inputs which occur somewhere in the training corpus, the ENTD-T is defined only for batch-aligned prefixes, which is a subset of valid inputs to the ENTD.

\subsection{Representing the ENTD and ENTD-T Distributions}
Naively computing the ENTD and ENTD-T distributions requires a full-corpus scan to compute the necessary counts. For a realistic modern LLM training corpus, this is computationally infeasible. We make use of $\infty$-gram \citep{liu2024infinigram}, a method which uses suffix arrays to efficiently count $k$-gram occurrences in large text corpora for arbitrary lengths $k$. A simple modification of $\infty$-gram also allows us to compute the batch-aligned ENTD-T counts efficiently.

\section{Experimental Setup}
\subsection{Models} \label{sec:models}
We use the Pythia model suite \citep{biderman2023pythia}. The models cover a wide range of model scales, from 70M to 12B parameters,\footnote{Pythia also contains several smaller models (as low as 14M parameters). We exclude some of the smaller models from the analysis since they clutter the figures and do not add much to our ultimate goal, which is to understand frontier models at scale.} and are trained on a large, realistic corpus (the Pile \citep{gao2020pile}). While these are still far from industrial scales, they are larger and more comprehensive than previous related studies \citep{nguyen2024ngram}. Importantly, Pythia contains model checkpoints throughout the training process, not just the final weights; and the \emph{dataloader} is reproducible, allowing us to know not only the full set of examples seen during the entirety of training, but specifically which samples were seen at which points. This will allow us to examine some of the effects of the training itself in detail.

We note an important caveat on the model suite. For the largest Pythia models (6.9B and 12B), the initialization parameters were not correctly specified; this is documented in the \href{https://github.com/EleutherAI/pythia/issues/135}{Pythia repo}. As such, it is possible that these models were trained sub-optimally, and at the very least there is a discrepancy between these two largest models and the remaining models which weakens our ability to directly compare them. We believe that these models can still give us valid insight into the data/model connection, but we will be more cautious about inferring scaling behavior, especially when the trends from the largest two models do not agree with the smaller models.

\subsection{Evaluation Data}
While there are many reasons why a transformer-based language model is preferred over the ENTD itself in practice, one of the critical ones for our setting is that the ENTD only makes predictions on inputs which are present in the training data. As such, we restrict ourselves to evaluation samples from the training set. While the ENTD can be extended to samples not contained in the training data via a backoff procedure (and other works, e.g., \cite{nguyen2024ngram}, take this approach), the backed-off distributions are not directly connected to the training loss in the same way that the ENTD is. Thus, we choose to sacrifice coverage of examples in order to retain a fully mathematically principled target.

If we simply took a uniformly random sample of examples used in training, the vast majority of samples used for our evaluations would occur only once in the dataset. This is because the Pythia models have a context window of length 2048 and a token vocabulary of order $10^5$. For inputs of even moderate length (e.g., 20 or more tokens), the odds of this string occurring multiple times in the corpus are slim, and any reoccurrences will tend to be duplicates of a longer string. In such cases, the ENTD and ENTD-T will be a point mass on the single completion seen for this exact string. A uniformly random sample from training examples would contain a fraction of at most $20/2048 \approx 1\%$ of examples outside of this regime, which does not give a good picture of the breadth of the model's behavior.

As such, we take an approach similar to \cite{carlini2022quantifyingmem} and define two stratified samples from the data. Specifically, we define several occurrence and length bins for the sampled training strings, and then uniformly sample an equal number of examples from each (occurrence count, length) bin. In some cases, the total number of samples in the corpus belonging to a given bin was below the desired count; in this case, we simply take all of the samples in the bin. Due to computational constraints, we shard the full Pile corpus, compute the stratified samples \emph{per-shard}, and then combine the samples across shards to get the final evaluation sample, removing duplicates that may have been chosen in multiple shards. We use this procedure to create one evaluation sample where individual examples can be pulled from anywhere in the corpus, and another evaluation sample where the examples are training context-aligned (and therefore valid for the ENTD-T). Statistics of the final samples are shown in Tables~\ref{tab:occupancy_ENTD} and \ref{tab:occupancy_ENTD-T}.

\section{Large Models Exhibit Strong Positional Invariance}

We first test the hypothesis which motivates the construction of the ENTD-T. Namely, the global minimizer of the loss that the model actually sees during training---where predictions are only made on batch-aligned prefixes---is the ENTD-T, not the ENTD. A natural conjecture is therefore that the model will approximately converge to the ENTD-T, rather than the ENTD.

\begin{figure}[t]
    \centering
    \includegraphics[width=\linewidth]{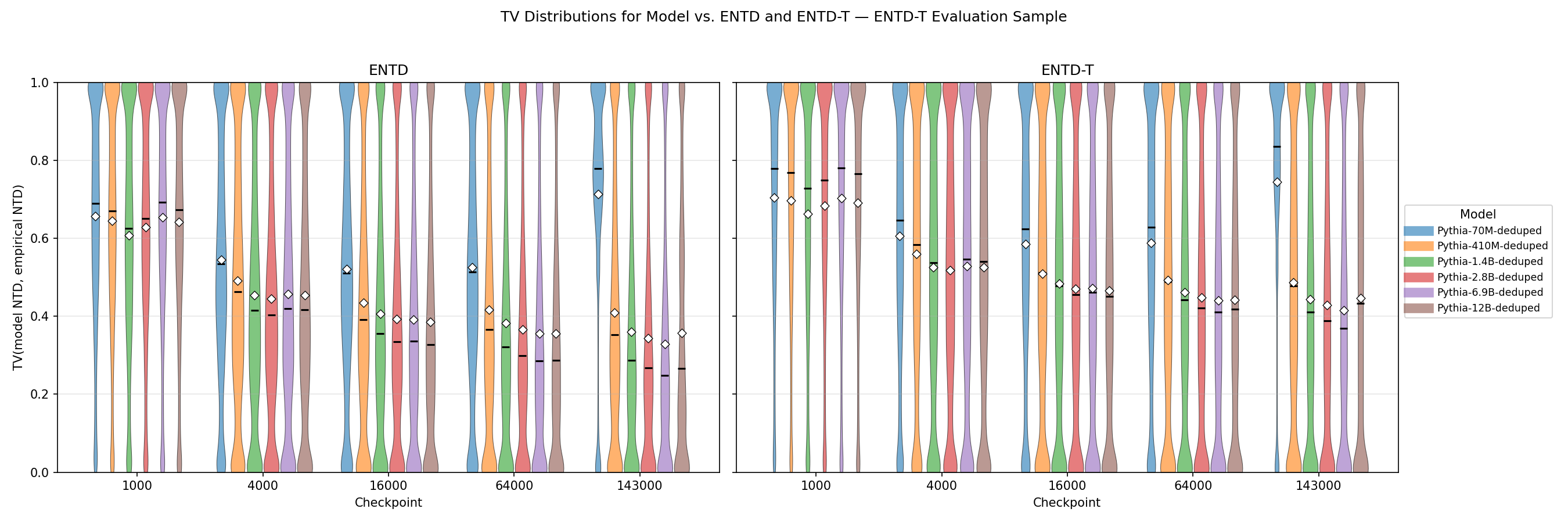}
    \caption{Distribution of discrepancies between the model and the ENTD (left) vs. ENTD-T (right). In spite of its more precise alignment with the training procedure, the ENTD-T generally has a higher discrepancy with the model's predictions as compared to the standard ENTD. In this and all other figures, the black bars denote medians and the white diamonds denote means.}
    \label{fig:model vs empirical violins}
\end{figure}

Figure~\ref{fig:model vs empirical violins} shows the distribution of TV discrepancies for the model vs. the ENTD (left) and vs. the ENTD-T (right). To provide a fair comparison, these discrepancies are computed on the ENTD-T evaluation sample (which are valid inputs to both the ENTD-T and ENTD). Surprisingly, even though the ENTD-T is aligned with the actual training procedure, the LLM distributions have consistently lower TV with the ENTD, with lower mean and median TV, and visually lighter tails at the upper end of the TV range.

There are two factors at play here. First, there is a sample efficiency issue: namely, because the ENTD-T only estimates empirical counts at batch-aligned positions, it has on average 1/(context window + 1) as many samples to estimate the empirical next-token distribution as compared to the ENTD. In the case of the Pythia models we used, this is a factor of 2049 sample discrepancy, which is significant. Since batch-aligned prefixes are \emph{statistically} identical to non-aligned samples, the increased sample size (and consequently reduced sampling error) of the ENTD vs. ENTD-T may outweigh the modest gains from batch alignment.

The second factor is that the conjectured batch alignment effect is much weaker than expected. Put another way, the transformer exhibits a \emph{strong positional invariance}, even though it was not explicitly trained to do so. This can be seen via the following experiment. The ENTD-T is constructed via the empirical counts for strings which are in positions that are at a 0-offset with respect to the training-aligned context windows. We can construct similar empirical distributions by simply changing the offset. We computed the TV distributions for the model vs. the empirical distributions with an offset of 1 (i.e., all strings which start one token after the training-aligned contexts, which gives slight misalignment) and 1024 (i.e., all strings which start 1024 tokens after the training contexts; this is in the exact middle of the training contexts, or maximally misaligned).

\begin{figure}[t]
    \centering
    \includegraphics[width=\textwidth]{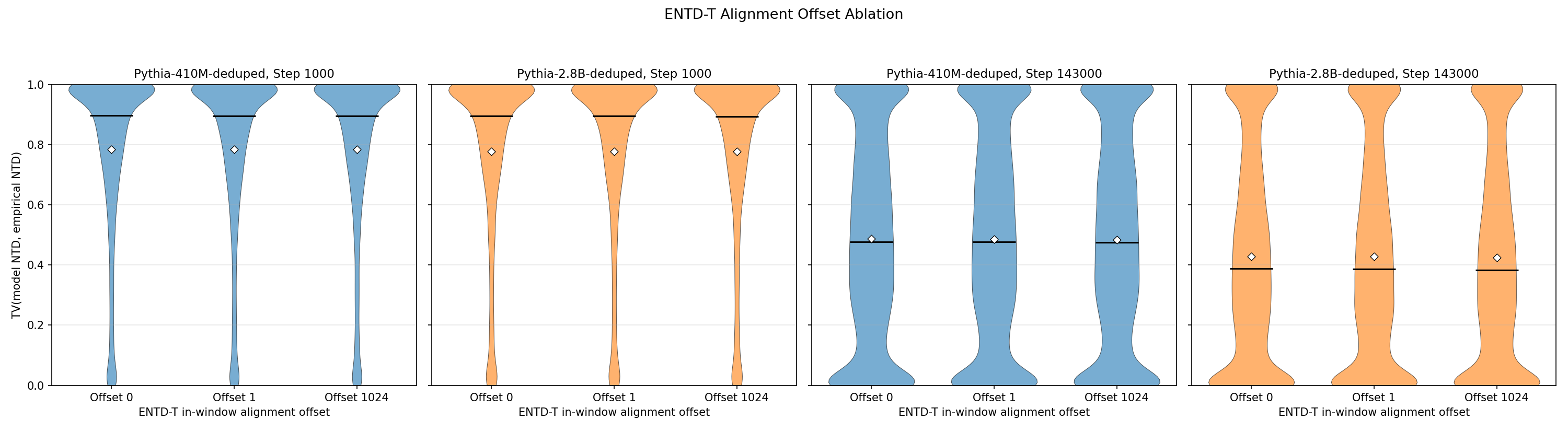}
    \caption{Discrepancies between the LLM and training-aligned empirical next-token distributions with different offsets; offset 0 corresponds exactly to the sequences seen in training. The LLM exhibits a strong positional invariance, as evidenced by the nearly identical discrepancy distributions across offsets.}
    \label{fig:offset invariance}
\end{figure}

The results are shown in Figure~\ref{fig:offset invariance}. Across different model sizes and throughout training, the TV distributions are nearly identical for different offsets. This indicates that the model has learned a strong positional invariance, in spite of the fact that the batch structure of training would not encourage this at first glance.

This is a surprising result. We therefore ran additional experiments to ensure that the positional invariance is real at these larger scales and not due to a bug in our empirical index construction. These results can be found in Appendix~\ref{app:ENTD-T confirmation}. In light of these results, all the remaining experiments will measure discrepancies between the LLM and the ENTD.

\section{The Softmax Bottleneck}
The final logits for a transformer are determined by dot products with $d_{\mathrm{model}}$-dimensional unembedding vectors, where $d_{\mathrm{model}}$ is the dimension of the model's residual stream. As such, it can be shown that for \emph{any} collection of $n$ inputs to the transformer, if the output logits for these examples are collected into a $n \times |V|$ matrix (where $|V|$ is the size of the token vocabulary; the matrix is formed by stacking the logit vectors in rows), then this matrix has rank at most $d_{\mathrm{model}}$. This is known as the softmax bottleneck \citep{yang2018softmaxbottleneck, carlini2024stealing}.

One possible explanation for the deviations between the LLM and empirical next-token distributions is that natural text does \emph{not} have a clean, low-dimensional structure, and indeed this hypothesis motivated \cite{yang2018softmaxbottleneck} to find ways around it. Nevertheless, the standard softmax is still the most popular choice for modern transformers, and this can perhaps explain more of the model/empirical discrepancy.

To test this, we can find an \emph{upper} bound on the bias introduced by the softmax bottleneck by finding a fully unconstrained rank-$d_{\mathrm{model}}$ logit approximation to the matrix of empirical distributions for the evaluation sample. Specifically, let $x^{(1)},\ldots,x^{(n)}$ be the set of evaluation strings and let $p_{\mathrm{ENTD}}(x^{(i)})$ be the empirical NTD for $x^{(i)}$, viewed as a vector in $\R^{|V|}$, and let $P \in \R^{n\times |V|}$ be the matrix with rows $p_{\mathrm{ENTD}}(x^{(i)})^\T$. Let $U\in \R^{n\times d}$ and $W\in \R^{|V|\times d}$. Here, the rows of $U$ can be thought of as the last layer token embeddings for the evaluation samples (before the unembedding), and $W$ stands in for the unembedding matrix. We then train $U$ and $W$ so that they minimize the cross-entropy loss: 
\[
\mathcal{L}(U, W) = \mathrm{CrossEntropy}(\mathrm{Softmax}(U W^\T), \: P),
\]
where the softmax is taken row-wise and the rows of $P$ are used as soft targets for the cross entropy. Essentially, this allows us to ask: if the \emph{only} constraint on the transformer's output distribution were due to the softmax bottleneck, and there were no constraints on the pre-softmax representations, how well could it approximate the empirical distribution?

\begin{figure}
    \centering
    \includegraphics[width=0.9\linewidth]{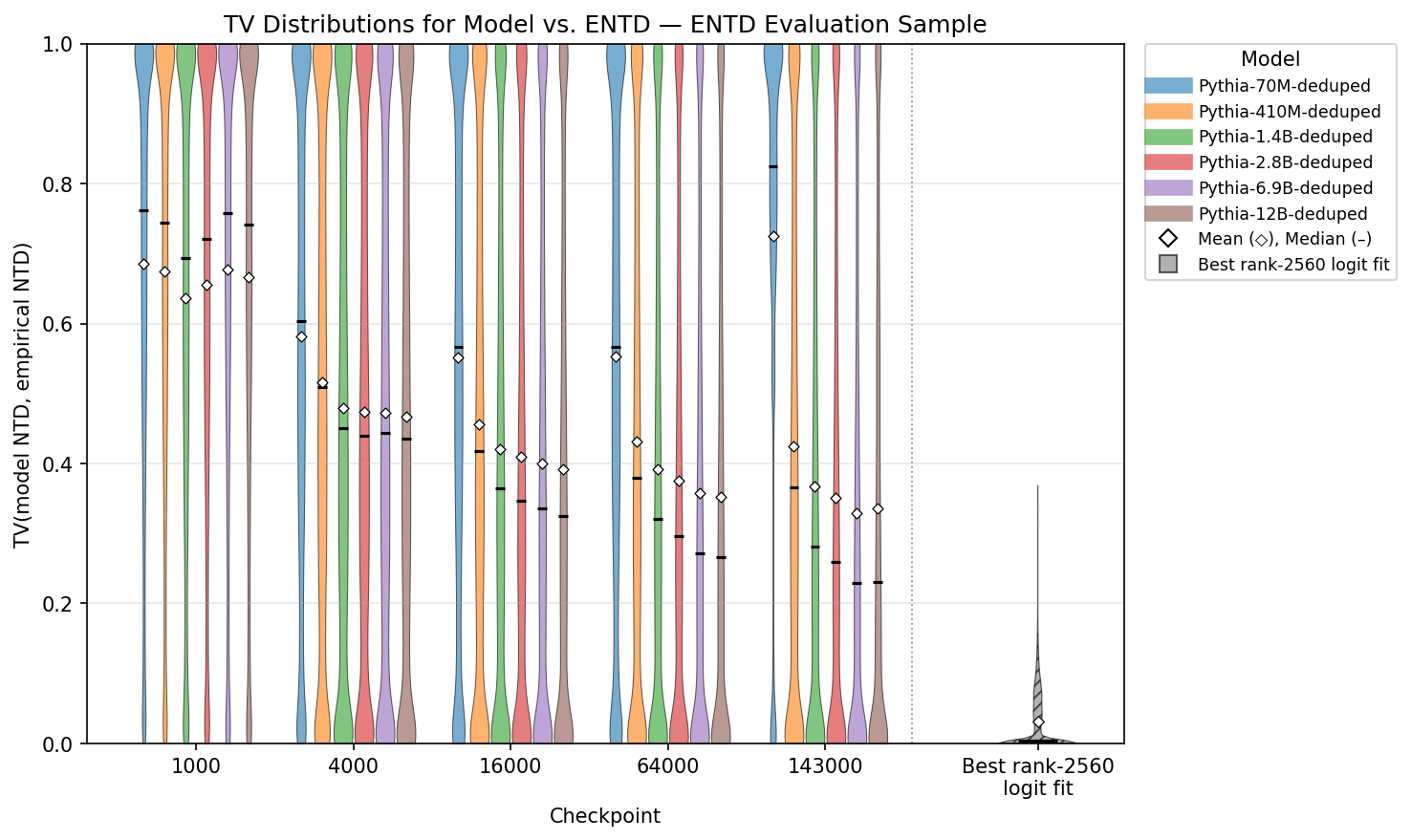}
    \caption{TV distributions on the ENTD evaluation sample for all models over training, plus the best rank-2560 ($d_{\mathrm{model}}$ for the 2.8B-parameter model) logit approximation to the evaluation sample distributions. Natural text (or at least, the subset of evaluation samples) is amenable to a low-rank approximation, but the transformer does not fully take advantage of this fact.}
    \label{fig:bottleneck}
\end{figure}

Surprisingly, we find that \cite{yang2018softmaxbottleneck}'s hypothesis does not hold: the empirical distributions of the evaluation samples are in fact highly amenable to a low-rank logit approximation, far below the deviation obtained by even the largest models at the end of training, shown in Figure~\ref{fig:bottleneck}. Thus, the natural empirical distributions (at least on our evaluation sample) \emph{are} amenable to a low-rank approximation, but the transformer does not take full advantage of this fact!

There is a caveat to this result, namely, that the best low-rank approximation on the evaluation sample (roughly 84K examples) is not an apples-to-apples comparison with how the transformer was trained, where it makes a total of $(\textrm{context length} = 2048) \times (\textrm{batch size} = 1024) \times (\textrm{training steps} = 143,000) \approx 300B$ predictions. The comparison may not be as unreasonable as it initially appears, however, because the number of samples used to build the full set of empirical NTDs on the evaluation sample (i.e., the sum of the number of corpus occurrences for each evaluation sample) is roughly 254B. Nevertheless, whether this low-rank structure persists as the number of evaluation samples grows is an open question.

\section{Optimization Randomness: Inter-model Discrepancy}
Training an LLM is an inherently stochastic procedure, with randomness arising from the ordering of the training data, random weight initialization, etc. It is therefore natural to ask, how much of the deviation between the LLM's predictions and the ENTD can be attributed to these random fluctuations, i.e., due to variance rather than a systemic bias?

\begin{figure}[t]
    \centering
    \includegraphics[width=\textwidth]{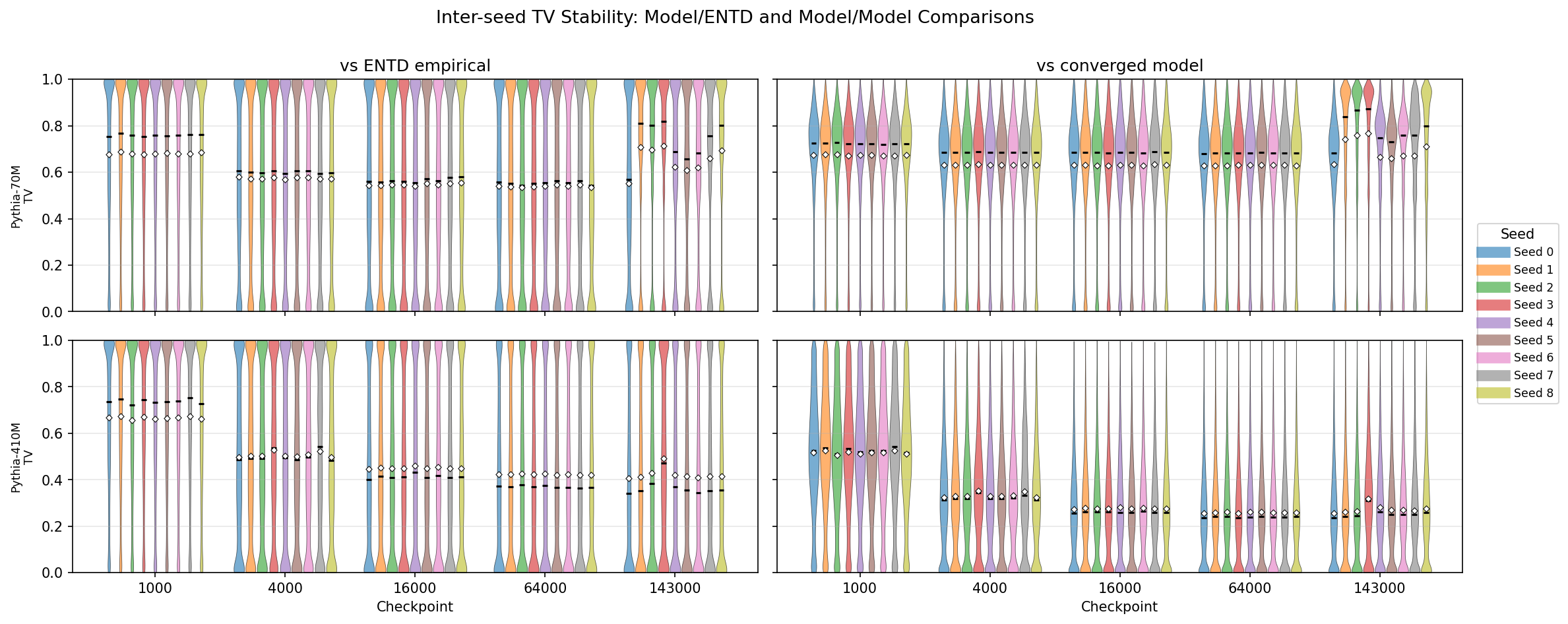}
    \caption{Model-ENTD deviations vs. model-model deviations. Discrepancies between each model and the ENTD are shown in the left column, while discrepancies between each model and a fixed, fully converged model of the same size are shown in the right column.}
    \label{fig:polypythias}
\end{figure}

To test this, we made use of the PolyPythias \citep{van2025polypythias} extension of the Pythia models. PolyPythias supplies model checkpoints from different random training runs using the same model architectures on the same overall corpus, but varying training order and random initializations. This allows us to directly measure the effects of randomness from the training process. The results are shown in Figure~\ref{fig:polypythias}, where we observe a number of interesting behaviors.

First, the shape of the model-empirical vs. model-model discrepancies are noticeably different: the model-empirical discrepancies tend to be bimodal at the extremes, whereas the model-model discrepancies generally have a modal outcome of moderate TV with a second mode of increasing size at very low TV values. This is consistent with the following picture: Given the training set and a fixed model size, the model learns an ``expected transformer distribution'' plus noise due to the random optimization. The shape of these discrepancy distributions is consistent with a fairly continuous/``well-behaved'' error distribution.

Another trend is that larger models converge to a more stable target, and the randomness from optimization has less of an impact. Indeed, for the 70M model, the average effect of the transformer bias (measured by the average discrepancy with the empirical distribution) is smaller than the model distribution variance (measured by average model-model discrepancy). But for the larger two models, the inter-model gap is smaller than the average gap between the models and the ENTD.

\section{Stratified Analysis}
\subsection{Different Lengths Are Learned Differently}

\begin{figure}[htbp]
    \centering
    \includegraphics[width=\linewidth]{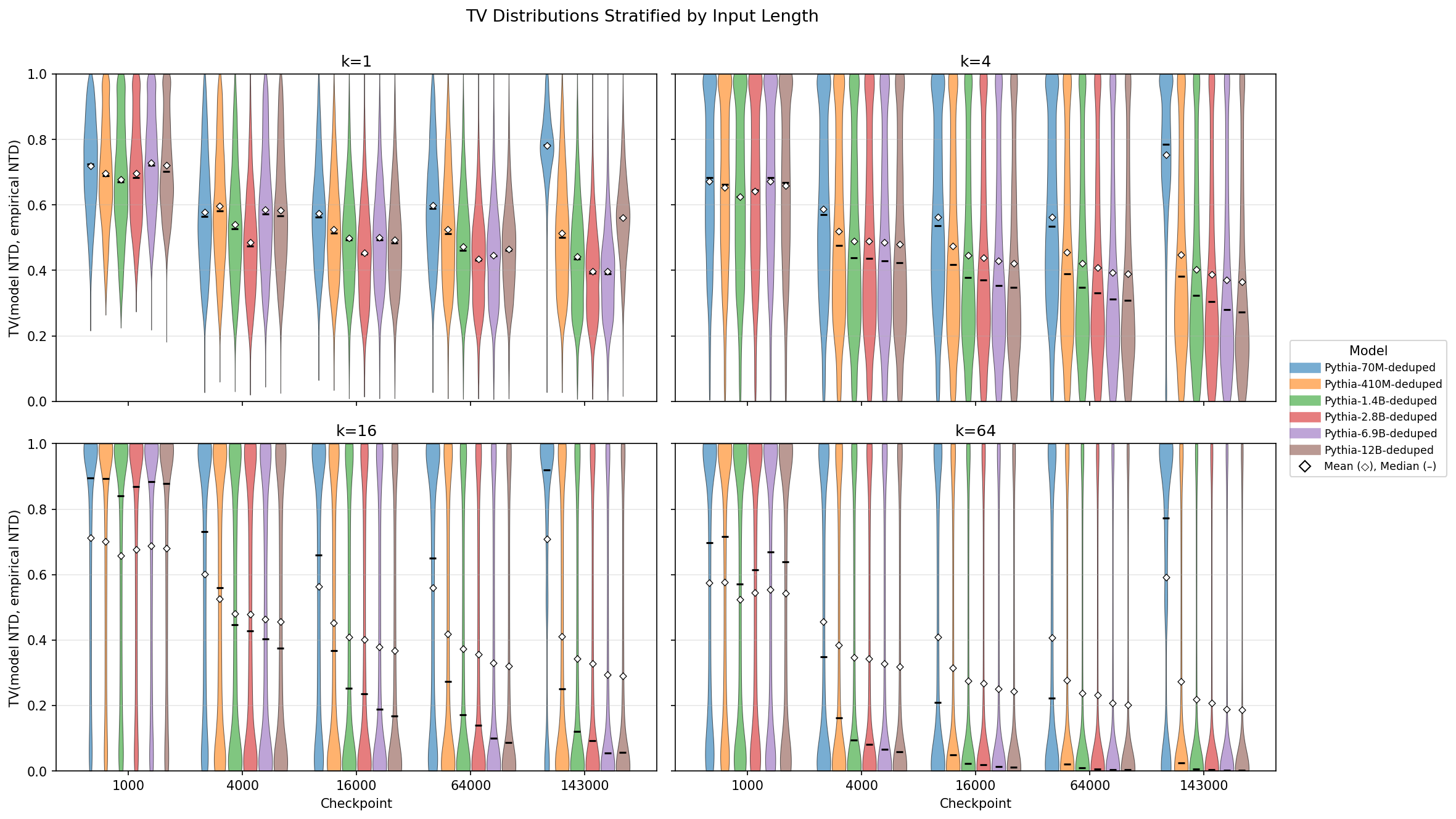}
    \caption{Model vs. ENTD discrepancies, stratified by evaluation sample length. The discrepancy distribution shapes vary greatly between input lengths. The general trend appears to be that the model expends less capacity to learn shorter inputs \emph{on average}, but longer inputs are more bifurcated, with some being learned perfectly while others are not learned at all. We also remark that these plots show that non-monotonic scaling of average TV for the 12B model is isolated entirely to the shortest $(k=1)$ sequences.}
    \label{fig:length stratified}
\end{figure}

Figure~\ref{fig:length stratified} shows TV distributions stratified by different input lengths (1, 4, 16, and 64) over the course of training. We observe a progressive ``stretching'' of the TV distributions from shorter to longer prefixes: the modal TV for shorter prefixes is moderate, whereas longer prefixes concentrate more and more at the extremes. One possible explanation for this is the relative importance of these terms to the overall training loss. Unigram discrepancy decreases moderately throughout training for larger models, but as these terms contribute fairly little to the overall loss, the model does not prioritize learning them exactly. In contrast, longer inputs contribute more to the loss, and the model expends more capacity to make many of these have very low discrepancy; but because there may be many challenging (or even ``bad'') long sequences, there is also a thicker tail of longer examples which the model does not learn at all.

\subsection{TV Tails}

For all but the smallest models, the modal outcome (widest part of the violin plot) is for the LLM to learn the ENTD almost perfectly (TV $\approx 0$). However, across model scales, there is a noticeable ``bump'' of evaluation samples with close to maximum discrepancy (TV = 1) with the ENTD, even at the end of training (refer to Figure~\ref{fig:bottleneck}). In this section, we examine whether there are any consistent properties which distinguish such samples from the rest of the data, which may explain why the model fails to learn them.

In Table~\ref{tab:tv tail stats}, we examine some aggregate statistics of evaluation samples in the most- and least-converged tails of the TV distributions, and compare them to the statistics of the overall evaluation sample. The general trends are consistent across model scales. For two of the statistics---input length and ENTD entropy---the high- and low-TV tails tend to lie at the extremes of the evaluation sample: both tails contain longer than average inputs and lower than average ENTD entropy. On the other hand, input frequency in the corpus and LLM entropy have more monotone relationships with TV, with median corpus frequency clearly decreasing with TV and entropy of the LLM's NTD clearly increasing with TV. The correlation with LLM entropy is sensible: when the model fails to fit the ENTD, it tends to be highly uncertain rather than confidently making an incorrect prediction.

\begin{table}[htbp]
\centering
\small
\setlength{\tabcolsep}{4pt}
\caption{Statistical comparison across TV-distance quantiles at step 143000 vs the full-corpus ENTD empirical, side by side for 410M-, 2.8B- and 12B-parameter models. ``Lo TV'' = bottom 10\% of per-prefix TV, ``Hi TV'' = top 10\%, ``All'' = all evaluation prefixes (identical set for every model, so non-model metrics repeat in the All columns). For each metric, the first row reports the mean and the second the median.}
\label{tab:tv tail stats}
\begin{tabularx}{\linewidth}{|>{\centering\arraybackslash}p{2.4cm}|CCC|CCC|CCC|}
\hline
\multirow{2}{*}{\textbf{Metric}} & \multicolumn{3}{c|}{\textbf{Pythia-410M}} & \multicolumn{3}{c|}{\textbf{Pythia-2.8B}} & \multicolumn{3}{c|}{\textbf{Pythia-12B}} \\
& \textbf{Lo TV} & \textbf{All} & \textbf{Hi TV} & \textbf{Lo TV} & \textbf{All} & \textbf{Hi TV} & \textbf{Lo TV} & \textbf{All} & \textbf{Hi TV} \\
\hline
\multirow{2}{*}{Length}
 & 32.82 & 14.65 & 17.31 & 33.53 & 14.65 & 16.52 & 32.37 & 14.65 & 16.13 \\
 & 32 & 8 & 12 & 32 & 8 & 12 & 32 & 8 & 8 \\
\hline
\multirow{2}{*}{Frequency}
 & 2.5e5 & 3.03e6 & 1,264 & 2.52e5 & 3.03e6 & 507 & 2.46e5 & 3.03e6 & 985 \\
 & 22,391 & 3,573 & 2 & 22,380 & 3,573 & 2 & 22,009 & 3,573 & 2 \\
\hline
\multirow{2}{*}{LLM Entropy}
 & 8.61e-3 & 3.14 & 4.81 & 2.00e-3 & 2.75 & 4.49 & 1.57e-3 & 2.86 & 4.45 \\
 & 5.07e-3 & 3.24 & 4.94 & 7.90e-4 & 2.64 & 4.61 & 6.91e-4 & 2.51 & 4.52 \\
\hline
\multirow{2}{*}{ENTD Entropy}
 & 4.67e-3 & 1.59 & 0.14 & 3.15e-3 & 1.59 & 0.20 & 2.58e-3 & 1.59 & 0.19 \\
 & 3.69e-3 & 0.24 & 0.00 & 2.99e-3 & 0.24 & 0.00 & 2.68e-3 & 0.24 & 0.00 \\
\hline
\end{tabularx}
\end{table}

We further examined individual examples in the TV tails and performed additional stratification to give a more fine-grained picture. Some examples are shown in Table~\ref{tab:tail examples Pythia-12B-deduped}; exact selection criteria are given in Appendix~\ref{app:selection}. Many of the examples in both the low- and high-TV tails are unsurprising. For instance, most of the low-TV examples are repeated many times in the corpus and have a deterministic completion, corresponding to, e.g., templates or code boilerplate. It is unsurprising that the model can learn these examples well. The consistently low median corpus count and ENTD entropy of the highest-TV tails also has a natural explanation: most inputs with high LLM/ENTD discrepancy can be attributed, in large part, to estimation error in the ENTD. These sequences have only one or two completions in the corpus, but the model does not overfit to these examples.

However, there is one surprising group of high-TV examples: those with a high corpus count (at least $10^4$) but which nevertheless have a high final TV.
We found that the high-TV, high-count examples tended to also have a high ENTD entropy; among high-count examples, the Spearman correlation between ENTD entropy and final TV was $+0.81$ for the 12B model. This is surprising: it means that it is difficult for the model to accurately represent genuine uncertainty in a way which is faithful to the true text distribution, even given many examples.

We also observed that the model does a better (though still imperfect) job of modeling high-entropy empirical distributions on longer inputs. Among high-count, high-entropy inputs, the Spearman correlation between the input length and the final TV was $-0.70$ for the 12B model. This is likely due to an inductive bias of the transformer architecture. For instance, it is known that chain of thought increases transformers' expressive power \citep{merrill2024expressive}; similarly, a longer input may in fact allow the transformer to express a larger class of next-token distributions.

\begin{table}[t]
\centering\small
\setlength{\tabcolsep}{8pt}
\caption{Lowest and highest final-TV evaluation inputs for Pythia-12B-deduped (TV between the model NTD and the full-corpus ENTD empirical at step 143{,}000). Count = corpus occurrences; $H_{\mathrm{ENTD}}$ and $H_{\mathrm{model}}$ = entropy (nats) of the ENTD and of the model NTD, respectively. Some inputs are truncated, denoted by ellipses (...).}
\label{tab:tail examples Pythia-12B-deduped}
\begin{tabular}{rrrrp{9.5cm}}
\toprule
\multicolumn{5}{l}{\textit{Best fit: deterministic templates / code / markup}}\\
\midrule
Count & TV & $H_{\mathrm{ENTD}}$ & $H_{\mathrm{model}}$ & Input Text \\
\midrule
7,205,946 & 0.00 & 0.00 & 0.00 & {\footnotesize\ttfamily .\textbackslash{}n\textbackslash{}nA:\textbackslash{}n} \\
6,402,143 & 0.00 & 0.00 & 0.00 & {\footnotesize\ttfamily ref-type=''supplementary-} \\
5,885,514 & 0.00 & 0.02 & 0.00 & {\footnotesize\ttfamily ?\textbackslash{}n\textbackslash{}nA:\textbackslash{}n} \\
5,380,615 & 0.01 & 0.52 & 0.53 & {\footnotesize\ttfamily -type=''table-fn} \\
\midrule\midrule
\multicolumn{5}{l}{\textit{Poor fit - High-entropy ENTD: frequent, short inputs with genuine next-token ambiguity}}\\
\midrule
Count & TV & $H_{\mathrm{ENTD}}$ & $H_{\mathrm{model}}$ & Input Text \\
\midrule
20,855 & 0.80 & 6.04 & 6.51 & {\footnotesize\ttfamily \textbackslash{}n\textbackslash{}t\textbackslash{}t\textbackslash{}t\textbackslash{}t\textbackslash{}t<h3>} \\
14,151 & 0.79 & 5.74 & 6.91 & {\footnotesize\ttfamily     @Expose\textbackslash{}n    private String} \\
18,562 & 0.77 & 3.86 & 6.88 & {\footnotesize\ttfamily       <type>} \\
33,697 & 0.76 & 5.70 & 6.53 & {\footnotesize\ttfamily  ISI\textbackslash{}n\textbackslash{}n} \\
\midrule\midrule
\multicolumn{5}{l}{\textit{Poor fit - Tiny-sample ENTD: model does not overfit to inputs which appear rarely}}\\
\midrule
Count & TV & $H_{\mathrm{ENTD}}$ & $H_{\mathrm{model}}$ & Input Text \\
\midrule
1 & 1.00 & 0.00 & 6.15 & {\footnotesize\ttfamily  there are some pretty serious challenges, like} \\
2 & 1.00 & 0.00 & 6.49 & {\footnotesize\ttfamily...to the instructions. By it's very nature the Rover} \\
2 & 1.00 & 0.00 & 5.35 & {\footnotesize\ttfamily  SiNx, Al2 O3 and the like and} \\
2 & 1.00 & 0.00 & 5.78 & {\footnotesize\ttfamily...effective start// time of other alarms to step past the}
 \\
\bottomrule
\end{tabular}
\end{table}

\begin{figure}
    \centering
    \includegraphics[width=0.75\linewidth]{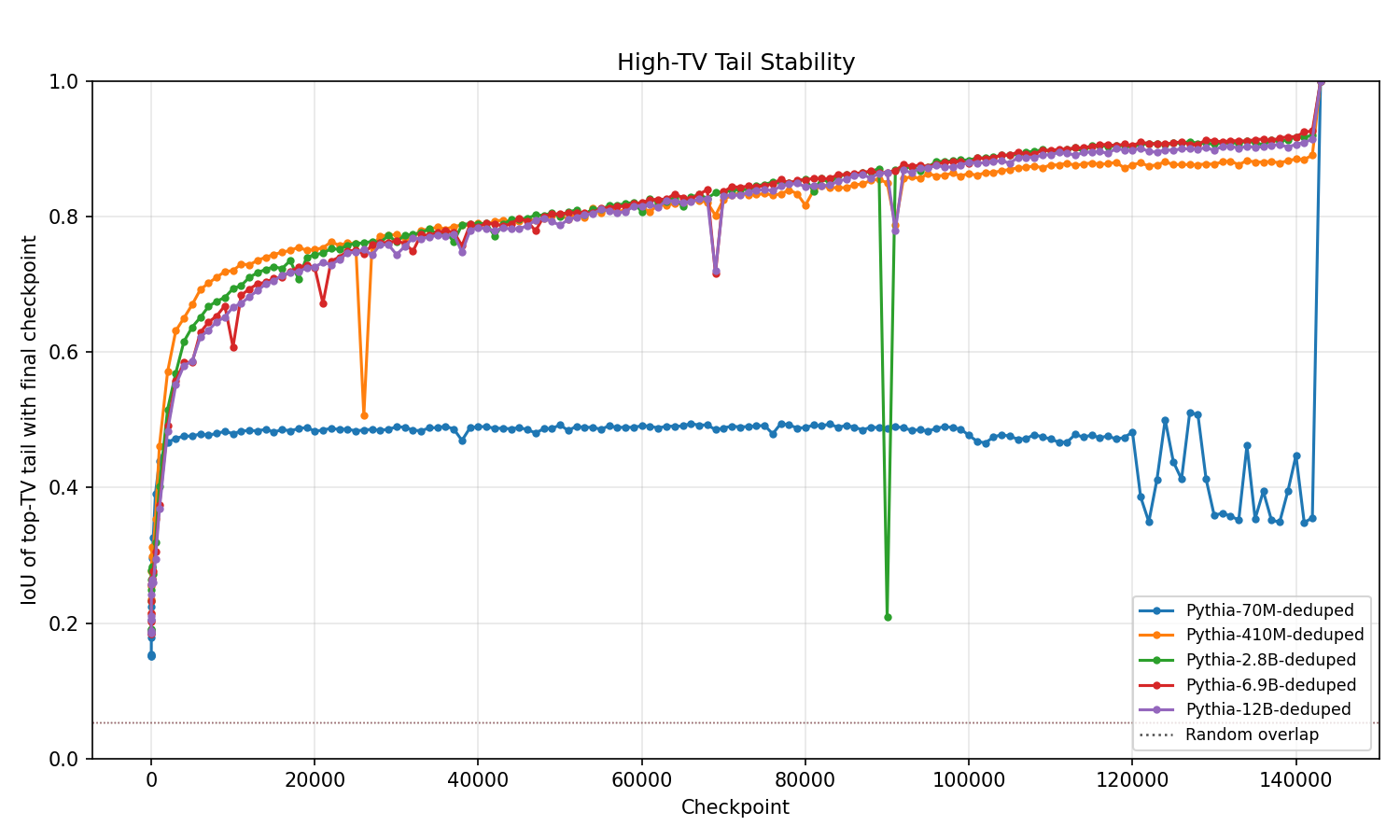}
    \caption{Stability of the high-TV tails, given by IoU of the top 10\% TV examples for each checkpoint with the top 10\% TV examples in the final checkpoint. The larger models exhibit high stability, indicating that there are a large number of structurally challenging examples for the model to learn.}
    \label{fig:tail stab}
\end{figure}

In addition to simply pattern matching on example strings from the tails, we also analyzed the stability of the tails over training to see whether or not these are ``structurally sound'' or just artifacts of a particular snapshot. The results are shown in Figure~\ref{fig:tail stab}, revealing that there is a combination of ``structurally challenging" examples which are consistently present, but a number of other examples which move in and out of the tail as the model learns.\footnote{We note briefly that the isolated dips for 410M and 2.8B correspond to single checkpoints where the model transiently collapses its next-token distribution onto a degenerate token in short contexts. For 12B, the small dips occur when the model's distribution transiently spreads towards uniform, also on small contexts. For these checkpoints, many of our evaluation samples temporarily have high TV, leading to a low IoU with the more stable final group. These events may be missed when only observing the aggregate training loss as these contexts account for a small fraction of the training data.} We also conducted a sensitivity analysis for the tail cutoff threshold and find qualitatively similar results. The comparison plots can be found in Appendix~\ref{app:sensitivity}. 

\section{Evolution of Individual NTDs}
While the distribution plots in Figure~\ref{fig:model vs empirical violins} give more information than a scaling law, they are still somewhat coarse. For instance, they cannot differentiate between the setting where the discrepancy for all examples slowly and steadily decreases relatively uniformly across examples (which would indicate that the model learns small amounts of generalizable information from each sample, and learns how to correctly respond on a given sample by aggregating this information across training); the setting where there are sudden decreases in a few samples at a time (which would indicate that the model learns by essentially memorizing one sample at a time); or some combination of the two or other more complicated patterns. In this section, we examine the change in TV for individual examples over the course of training to shed light on this picture.

\begin{figure}[t]
    \centering
    \includegraphics[width=0.49\textwidth]{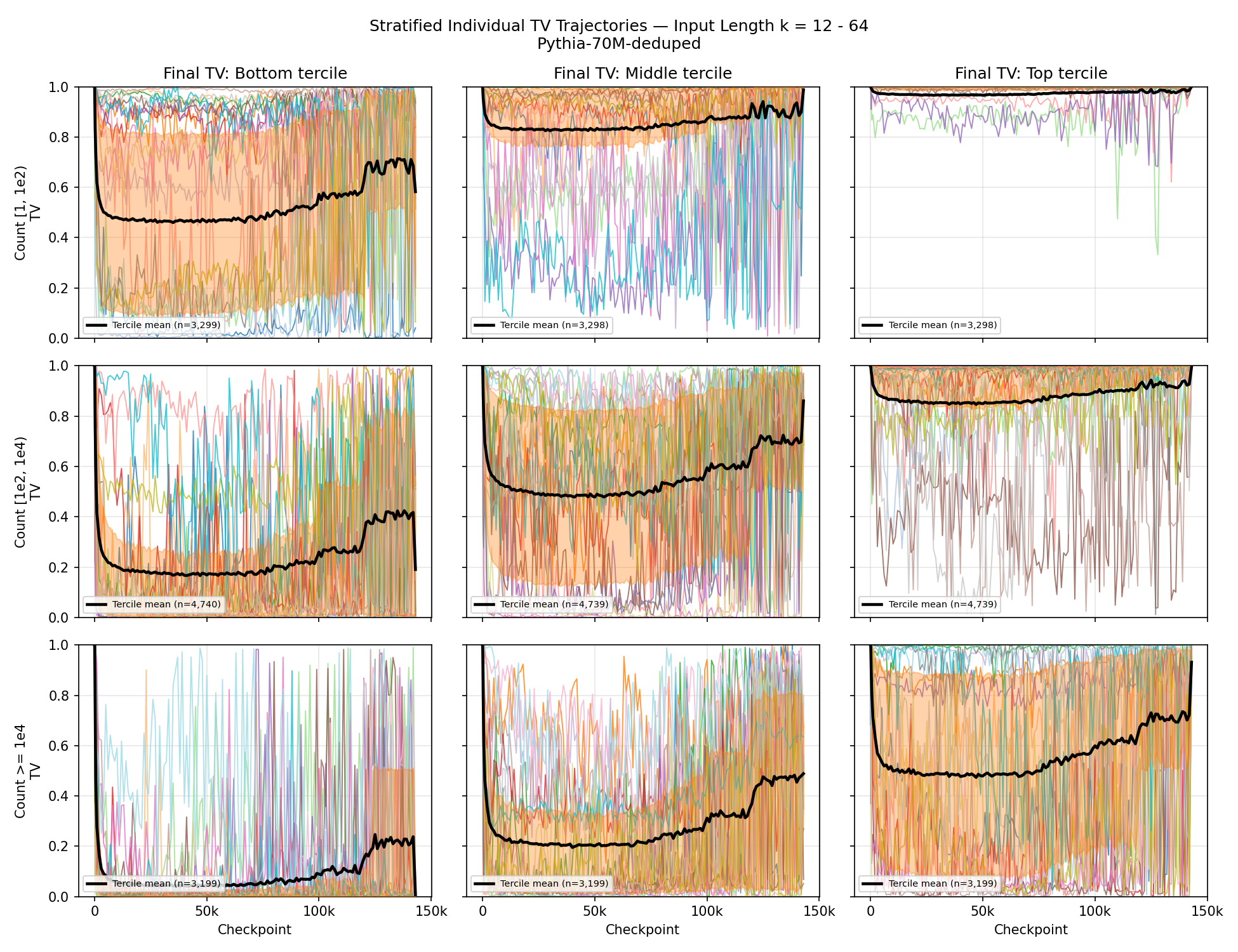}
    \includegraphics[width=0.49\textwidth]{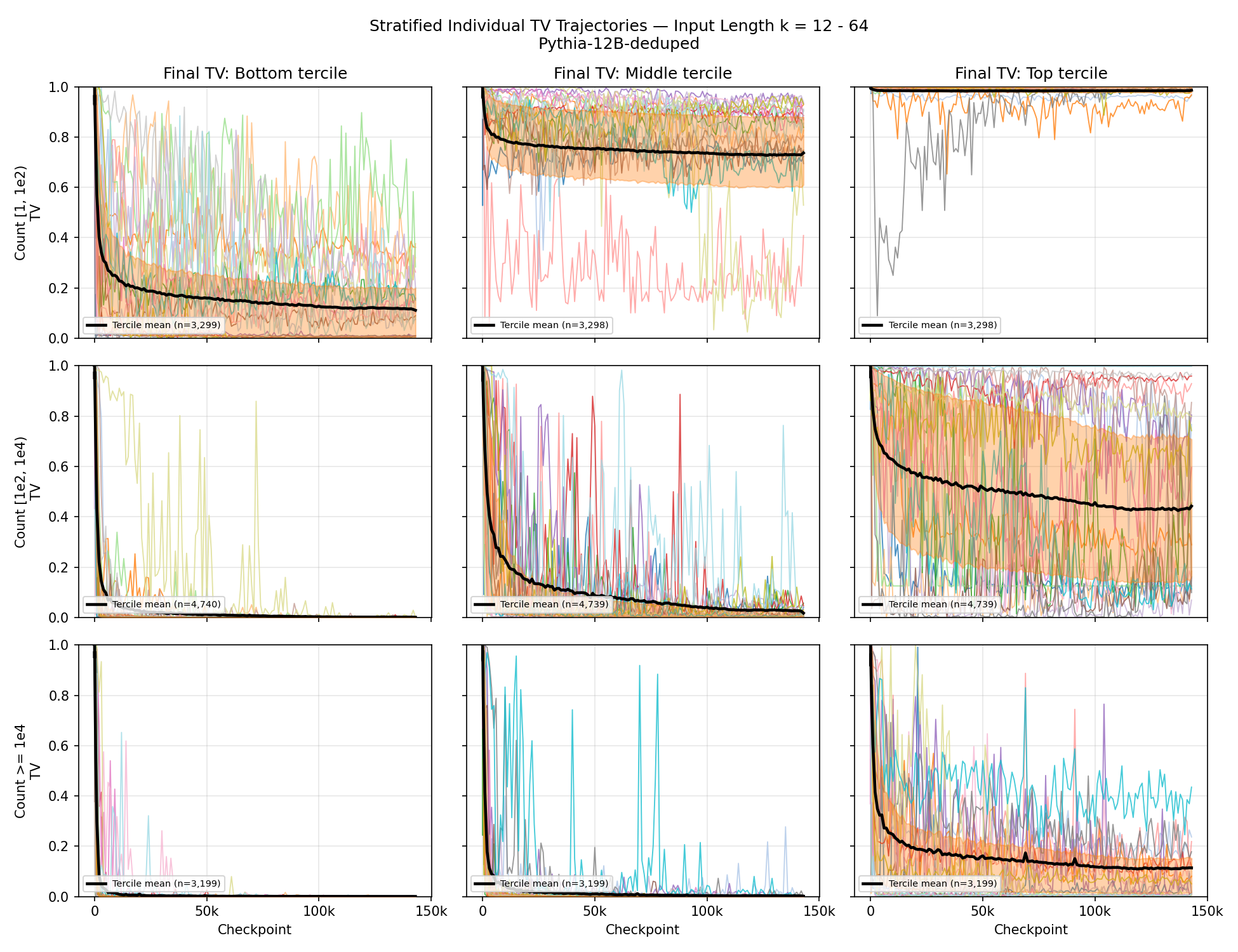}
    \caption{Individual TV trajectories for a small (70M, left) and large (12B, right) model. The trajectories exhibit much greater oscillation in the smaller model, aligning with the findings of other recent work \citep{huang2026larger}. For the larger model, examples at the extreme ends of the discrepancy distribution (TV$\approx$ 0 or 1) tended to have more stable trajectories, while more oscillation is observed in the middle of the TV range.}
    \label{fig:individual}
\end{figure}

Results stratified by evaluation example count and its TV at the final training step are shown in Figure~\ref{fig:individual} for a small model (70M, left) and a large model (12B, right). As expected for such a complicated system, the individual trajectories are a combination of many distinct behaviors, but we observe a few general trends.

First, the TV trajectories for the small model are in general much more oscillatory than they are for the large model. This accords with the recent work of \cite{huang2026larger}, who showed that gradients of distinct samples exhibit much greater interference in smaller models.

For the large model, there are two categories of points which have exceptionally oscillatory TV trajectories: inputs with low counts but which nevertheless have low to medium final TV; and inputs with higher counts but with high final TV. In the former case, these low-count and low/medium final TV examples tend to have a point-mass ENTD (i.e., a single completion in the corpus) but are nevertheless predictable, for instance, code stack traces or enumeration formats. This has two consequences. First, because the ENTD is a point mass, the TV curve is just $1-p_t(c)$, where $p_t(c)$ is how much mass the model places on the single completion $c$ at step $t$. Since this is only dependent on a single token probability, there is no averaging effect over tokens, making it easier for variation to occur. Second, because these exact sequences are rare, it is unlikely that the model can memorize them exactly; but because they are predictable, generalizable information from other sequences keeps $p_t(c)$ fluctuating in a medium range. This is in contrast to the low-count but high-TV examples, which have low variability: the model does not memorize these either, but their single observed completion is not strongly predictable from related text. Even when the input is perfectly natural, the observed continuation is only one of many plausible ones, so the model's mass on it stays near zero, where fluctuations are necessarily negligible and the TV remains pinned close to 1.


\section{Discussion}
In this work, we studied the extent to which LLM predictions can be explained by the empirical next-token distribution of the data, which is the global minimizer of the training loss and therefore ostensibly what training should push the model towards. While this does indeed occur \emph{on average} over the training samples---a fact which is essentially equivalent to the existence of neural scaling laws---a more fine-grained look at the discrepancies of \emph{individual samples} reveals that while the empirical distribution models many inputs well, there is also a long tail of examples which it does not. We identified several aspects of the LLM stack which could contribute to discrepancies between the LLM and the ENTD, including training context window alignment, randomness in the optimization procedure, and representational limitations of transformers.

We found that transformers exhibit a strong positional invariance, negating the batch-alignment effect.
We also found that the empirical distribution of natural data seems to be amenable to low-rank logit approximation, and the learned transformer exhibits stronger smoothing of the empirical distribution than what could be attributed solely to the softmax bottleneck effect.
The randomness of the optimization procedure also seems to have some role in the deviation, but it becomes less pronounced as the model grows in scale.
Finally, the LLM surprisingly struggles to model genuinely uncertain next-token distributions even given many training examples, though this effect is reduced on longer inputs.

\paragraph{Limitations and Future Work}
These models are below frontier scale, and there is always a risk that our conclusions will not hold or will be qualitatively dissimilar at the frontier. Nevertheless, we have some confidence that our conclusions are relevant given that they hold across more than two orders of magnitude model parameters. There is an additional issue (noted in the \href{https://github.com/EleutherAI/pythia/issues/135}{Pythia repo}): for the largest Pythia models (6.9B and 12B), the initialization parameters were not correctly specified, meaning these largest models may have been trained sub-optimally. This may account for some of the plateaus or inconsistencies noted in the scaling behavior, e.g. in Figs.~\ref{fig:model vs empirical violins} and \ref{fig:bottleneck}.

Another key limitation is that our strict ENTD analysis is only applicable to sequences belonging to the training set. The $\infty$-gram model can be extended to sequences not seen during training (indeed, the authors provide this functionality in the official release via backoffs to substrings of the input which are contained in the training set, and other works such as \cite{nguyen2024ngram} use these backoffs to analyze strings not belonging to the training set). However, as the backoff procedure has no principled connection to the model's training, we chose to limit our scope to training examples only, rather than weaken the connection between the LLM and the object we are using to explain it. Nevertheless, explaining how the model produces its outputs on \emph{unseen} inputs is clearly a desirable goal for future work.

An orthogonal issue is that our analysis focuses on the model's predicted distribution for a single next token. However, LLMs' usefulness comes not only from predicting one single token into the future, but rather from chaining these distributions together to \emph{generate} long sequences of new tokens. As the NTD is the atomic component of these generations, and this much simpler object is still far from completely understood, we chose to focus on the NTD in this work. However, using these atomic results to say something meaningful about the long generations of an LLM seems challenging in its own right.

\section{Related Work}
There have been numerous attempts to draw a more precise link between LLMs and their training data. Some works have attempted to trace model behavior to \emph{individual} datapoints using approximate methods such as influence functions \citep{grosse2023influence, mlodozeniec2026distributionalinfluence, chen2026mechanisticinfluence} or by finding training sequences which match part of the output \citep{liu2025olmotrace}. In contrast, our work considers \emph{distributional} properties of the data, rather than individual points. Previous work suggests that some model behaviors may not be attributable to individual points, but rather aggregate statistics of many data (e.g. \cite{izzo2026quantitative}).

Other works have tested how training data statistics impact a model's performance on specific tasks or behaviors \citep{wei2021frequency, elazar2022factual, mccoy2024embers, allen2024physics, razeghi2024backtracking, chan2026entitycomparison} or on heuristic measures of generalization \citep{wang2025tracing}. \cite{im2026associate} studied the connection between training data statistics and the internals/representations learned by a transformer (rather than the final output distribution), focusing on early stages in training.

The work most similar in spirit to ours is \cite{nguyen2024ngram}, which also explored the agreement between LLM predictions and $n$-gram-based models (such as the ENTD). However, there are several important differences.
Most of their experiments use a 160M-parameter transformer on TinyStories, which contains 480M tokens. The largest tested model has 1.4B parameters on a 4.4B-token training set. They include experiments which measure the TV distance between the model output and the ENTD for 7-gram sub-sequences of the input and for the fully trained model.
In contrast, we explore a much wider range of model sizes/compute scales, as well as a wider range of $n$-grams, and more in-depth discussion of where the discrepancy between the model and the ENTD may emerge from.

In summary, with a few exceptions, most of these studies have proceeded via heuristic measurements of the data or by interrogating the models with specific, human-intelligible tasks. Our approach attempts to take the bare mathematical basis of LLM pretraining---training the model to perform next-token prediction via the cross-entropy loss---at face value to determine how far these quantitative facts can take us.

\bibliographystyle{tmlr}
\bibliography{main}

\newpage
\appendix

\section{Evaluation Sample Statistics} \label{app:eval sample}

\begin{table}[htbp]
\centering
\small
\setlength{\tabcolsep}{4pt}
\caption{Corpus-level occupancy of the ENTD stratified evaluation sample: number of prefixes per (length $k$, corpus-count bin), with counts taken as all-position (ENTD) occurrences. Reservoir size $N=100$ per $(k,\text{shard-bin})$ per shard (21 shards); each prefix is re-binned by its corpus-wide count in the combine phase. Total 83,963 prefixes.}
\label{tab:occupancy_ENTD}
\begin{tabular}{lrrrrrrrr}
\hline
$k$ & $[10^{0},\, 10^{1})$ & $[10^{1},\, 10^{2})$ & $[10^{2},\, 10^{3})$ & $[10^{3},\, 10^{4})$ & $[10^{4},\, 10^{5})$ & $[10^{5},\, 10^{6})$ & $[10^{6},\, \infty)$ & \textbf{Total} \\
\hline
1 & 2 & 0 & 18 & 75 & 503 & 2,102 & 3,761 & 6,461 \\
2 & 560 & 1,247 & 1,887 & 2,050 & 2,142 & 2,169 & 2,773 & 12,828 \\
4 & 1,405 & 726 & 1,789 & 2,031 & 2,051 & 2,038 & 1,125 & 11,165 \\
6 & 1,864 & 484 & 1,671 & 1,983 & 2,045 & 1,761 & 329 & 10,137 \\
8 & 1,985 & 485 & 1,589 & 1,953 & 2,036 & 1,411 & 203 & 9,662 \\
12 & 2,040 & 489 & 1,549 & 1,942 & 1,986 & 834 & 110 & 8,950 \\
16 & 2,064 & 482 & 1,547 & 1,958 & 1,935 & 532 & 77 & 8,595 \\
32 & 2,085 & 365 & 1,621 & 1,965 & 1,912 & 273 & 52 & 8,273 \\
64 & 2,080 & 290 & 1,675 & 1,961 & 1,736 & 136 & 14 & 7,892 \\
\hline
\textbf{Total} & \textbf{14,085} & \textbf{4,568} & \textbf{13,346} & \textbf{15,918} & \textbf{16,346} & \textbf{11,256} & \textbf{8,444} & \textbf{83,963} \\
\hline
\end{tabular}
\end{table}

\begin{table}[htbp]
\centering
\small
\setlength{\tabcolsep}{4pt}
\caption{Corpus-level occupancy of the ENTD-T stratified evaluation sample: number of prefixes per (length $k$, corpus-count bin), with counts taken as batch-aligned (ENTD-T) occurrences. Reservoir size $N=100$ per $(k,\text{shard-bin})$ per shard (21 shards); each prefix is re-binned by its corpus-wide count in the combine phase. Total 56,052 prefixes.}
\label{tab:occupancy_ENTD-T}
\begin{tabular}{lrrrrrrrr}
\hline
$k$ & $[1,\, 3)$ & $[3,\, 10)$ & $[10,\, 30)$ & $[30,\, 100)$ & $[100,\, 300)$ & $[300,\, 1,000)$ & $[1,000,\, \infty)$ & \textbf{Total} \\
\hline
1 & 5 & 70 & 197 & 772 & 2,407 & 2,312 & 4,542 & 10,305 \\
2 & 764 & 659 & 743 & 1,355 & 1,734 & 1,931 & 3,604 & 10,790 \\
4 & 1,774 & 313 & 580 & 1,152 & 1,492 & 1,585 & 854 & 7,750 \\
6 & 1,997 & 188 & 601 & 1,023 & 1,269 & 737 & 192 & 6,007 \\
8 & 2,065 & 116 & 509 & 1,081 & 1,009 & 369 & 94 & 5,243 \\
12 & 2,077 & 131 & 583 & 985 & 490 & 115 & 67 & 4,448 \\
16 & 2,086 & 157 & 611 & 899 & 261 & 89 & 63 & 4,166 \\
32 & 2,097 & 159 & 714 & 639 & 130 & 45 & 44 & 3,828 \\
64 & 2,096 & 205 & 703 & 435 & 60 & 5 & 11 & 3,515 \\
\hline
\textbf{Total} & \textbf{14,961} & \textbf{1,998} & \textbf{5,241} & \textbf{8,341} & \textbf{8,852} & \textbf{7,188} & \textbf{9,471} & \textbf{56,052} \\
\hline
\end{tabular}
\end{table}

\section{Confirmation of the Null Batch Alignment Result} \label{app:ENTD-T confirmation}

\begin{figure}[ht!]
    \centering
    \includegraphics[width=0.49\textwidth]{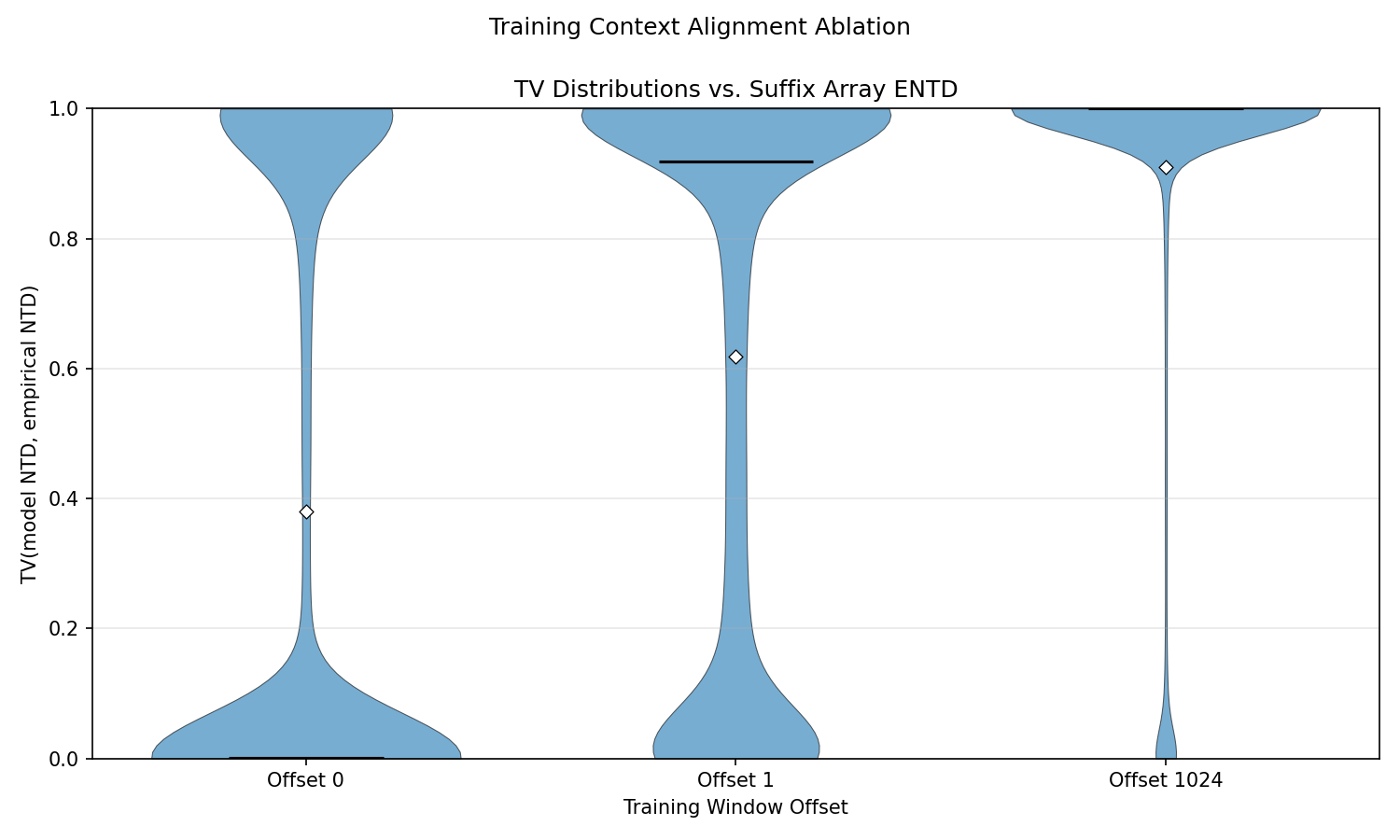}
    \includegraphics[width=0.49\textwidth]{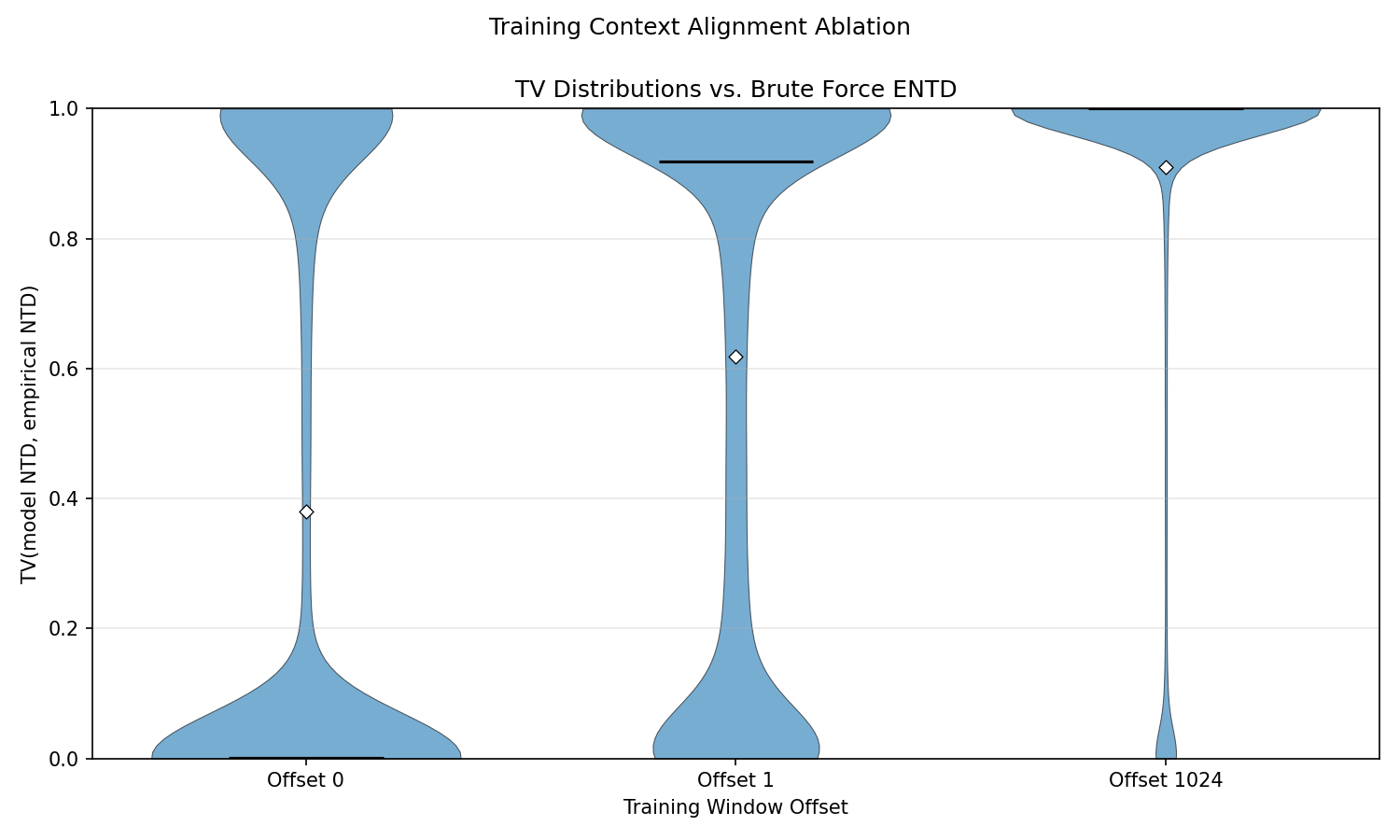}
    \caption{Batch-aligned vs. offset discrepancies after extended training on only two batches, and with the empirical indices computed via the infini-gram approach used in the other experiments (left) as well as a brute-force (but easily verifiably error-free) method for constructing the empirical indices (right). When the model is trained extensively on only two batches, the expected batch-aligned discrepancy gap emerges. The infini-gram-based indices also align perfectly with the brute-force method.}
    \label{fig:ENTD-T confirmation}
\end{figure}

Figure~\ref{fig:ENTD-T confirmation} shows the discrepancies when the model is trained for 1000 steps on two batches (500 steps on each batch). In this case, we get the expected result, namely, the model has overfit only to the batch-aligned sequences. Even with a single token offset, the gap widens significantly, and with an offset of 1024 tokens (exactly half the context window size) the median gap is close to maximal.

We also constructed the empirical next-token distributions via two methods: the infini-gram-based approach used for the rest of the experiments, as well as a ``brute force'' method, where we simply kept a hash table of next tokens following inputs seen during training. Notably, the brute-force index was constructed directly from the logged and independently verified training data order, without a suffix array, allowing us to directly verify that the sequences being fed to the model (or with the appropriate offset) are in fact the same as the ones being used to construct the index. In addition to the perfectly overlapping TV distributions in Figure~\ref{fig:ENTD-T confirmation}, we also confirmed that the empirical indices constructed via these two approaches agreed exactly on all entries.

\section{Selection Criteria for Examples in Table~\ref{tab:tail examples Pythia-12B-deduped}} \label{app:selection}

We used the following rules to select the specific examples in Table~\ref{tab:tail examples Pythia-12B-deduped}. In each block we additionally skip any input that shares a $\ge 20$-character substring with an already-selected input, so that the table does not show near-duplicate windows of the same underlying text.

\emph{Best fit}: We first filter to samples with $\mathrm{TV} \le 0.03$ and $k \ge 6$. From among these samples, we select the four with the highest corpus count. 

\emph{High-entropy ENTD}: We first filter to samples with count $\ge 10^{4}$, empirical entropy $H_{\mathrm{ENTD}} \ge 3$, and length $k \ge 4$. From among these samples, we choose the four examples with the highest TV.

\emph{Tiny-sample empirical}: We filter to samples with count $\le 2$, length $8 \le k \le 32$, and final $\mathrm{TV} \ge 0.995$. Because every input in this pool lies within $0.005$ TV of the maximum, we draw the examples uniformly at random with a fixed seed.

\section{Tail Stability Sensitivity Analysis} \label{app:sensitivity}

\begin{figure}[ht!]
    \centering
    \includegraphics[width=\linewidth]{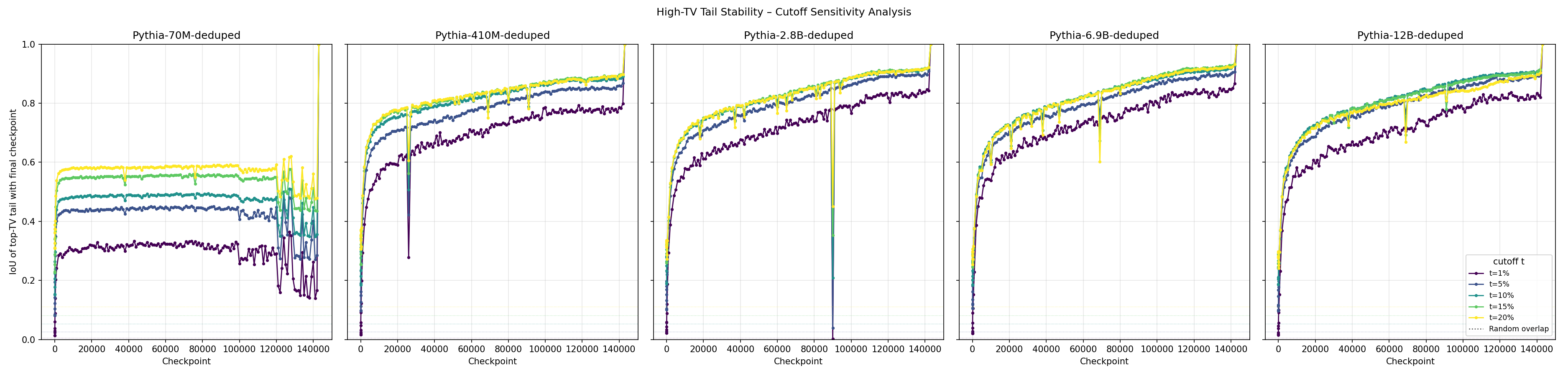}
    \caption{Tail stability analysis. The shape of the tail stability curve is stable across choices of the cutoff. The isolated single-checkpoint dips are discussed in Figure~\ref{fig:tail stab}.}
    \label{fig:tail stab sensitivity}
\end{figure}

The qualitative tail stability picture is fairly robust to the chosen cutoff. Refer to the plots in Figure~\ref{fig:tail stab sensitivity}. The shape of the curves is stable across thresholds.

\newpage

\section{Additional Trajectories} \label{app:trajectories}

\begin{figure}[ht!]
    \centering
    \includegraphics[width=0.49\textwidth]{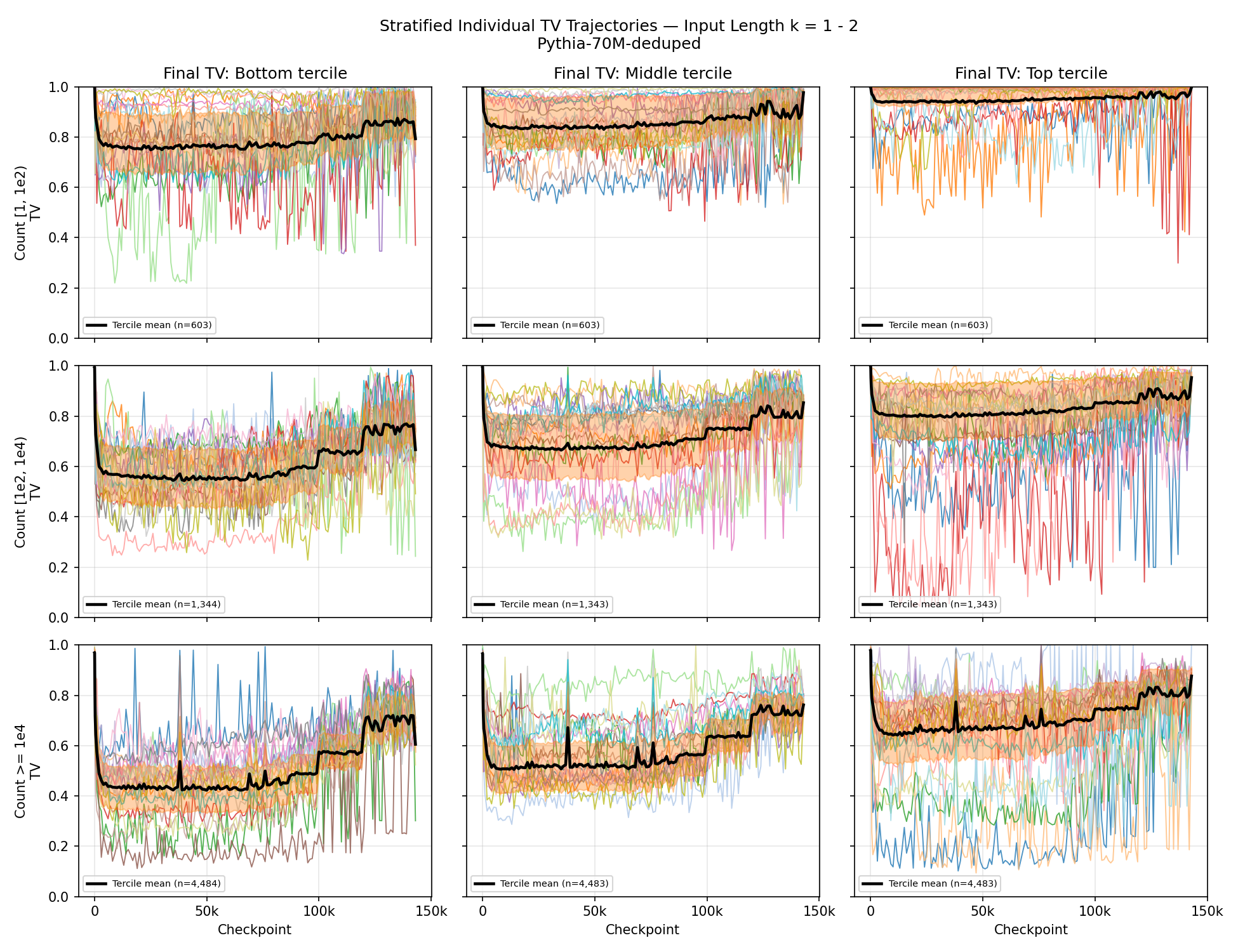}
    \includegraphics[width=0.49\textwidth]{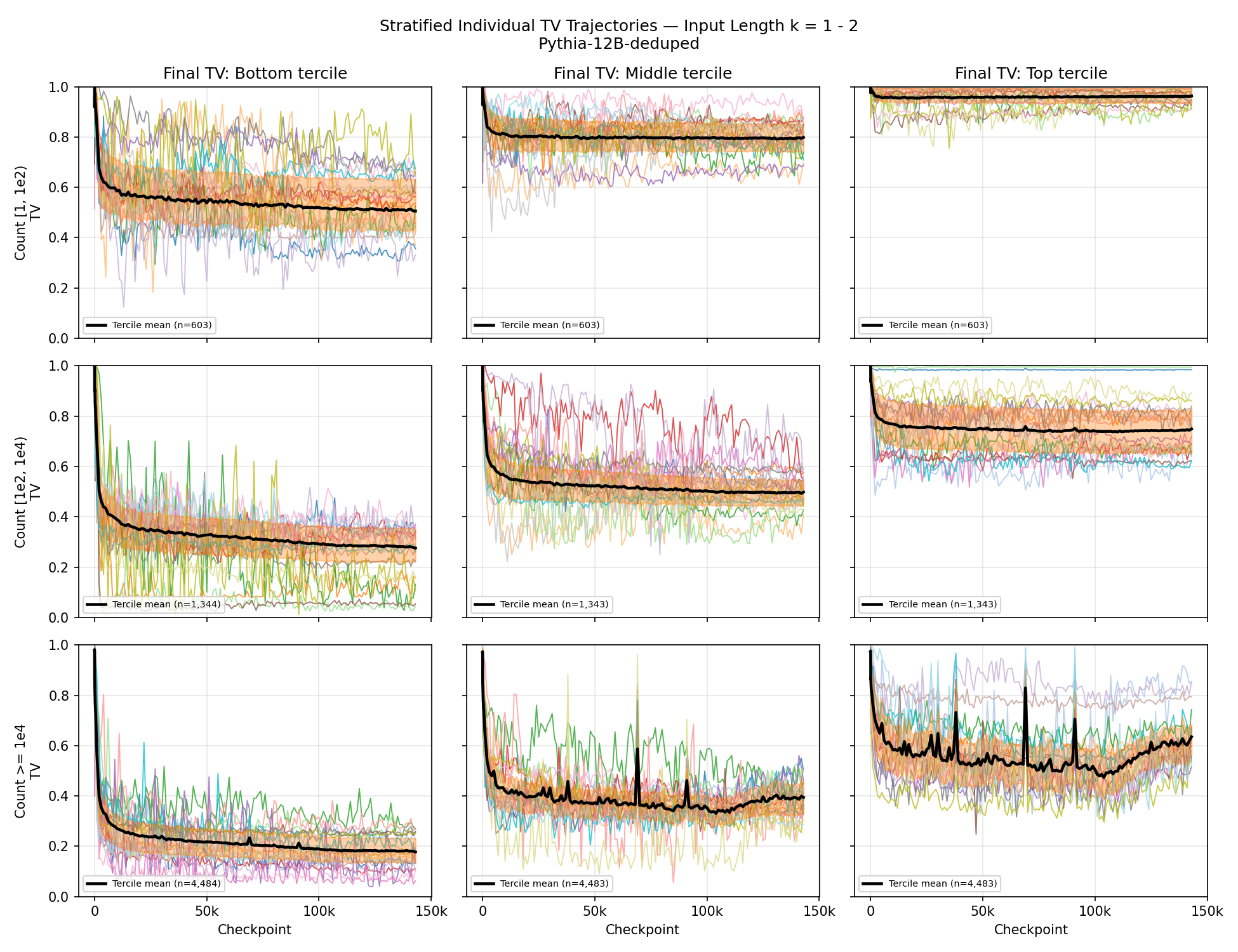}
    \caption{Individual TV trajectories for a small (left) and large (right) model, input length $k=1-2$.}
    \label{fig:individual k1-2}
\end{figure}

\begin{figure}[ht!]
    \centering
    \includegraphics[width=0.49\textwidth]{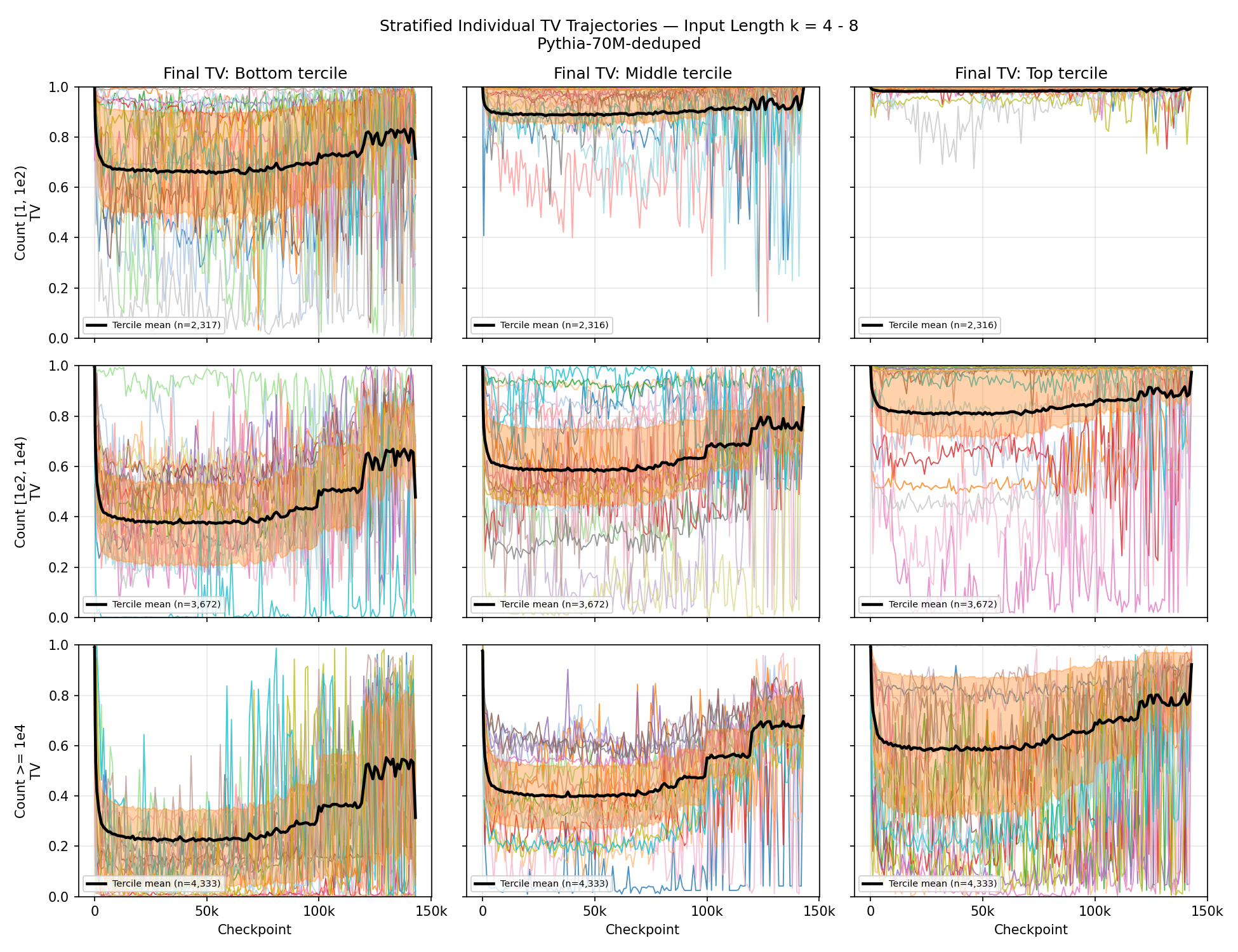}
    \includegraphics[width=0.49\textwidth]{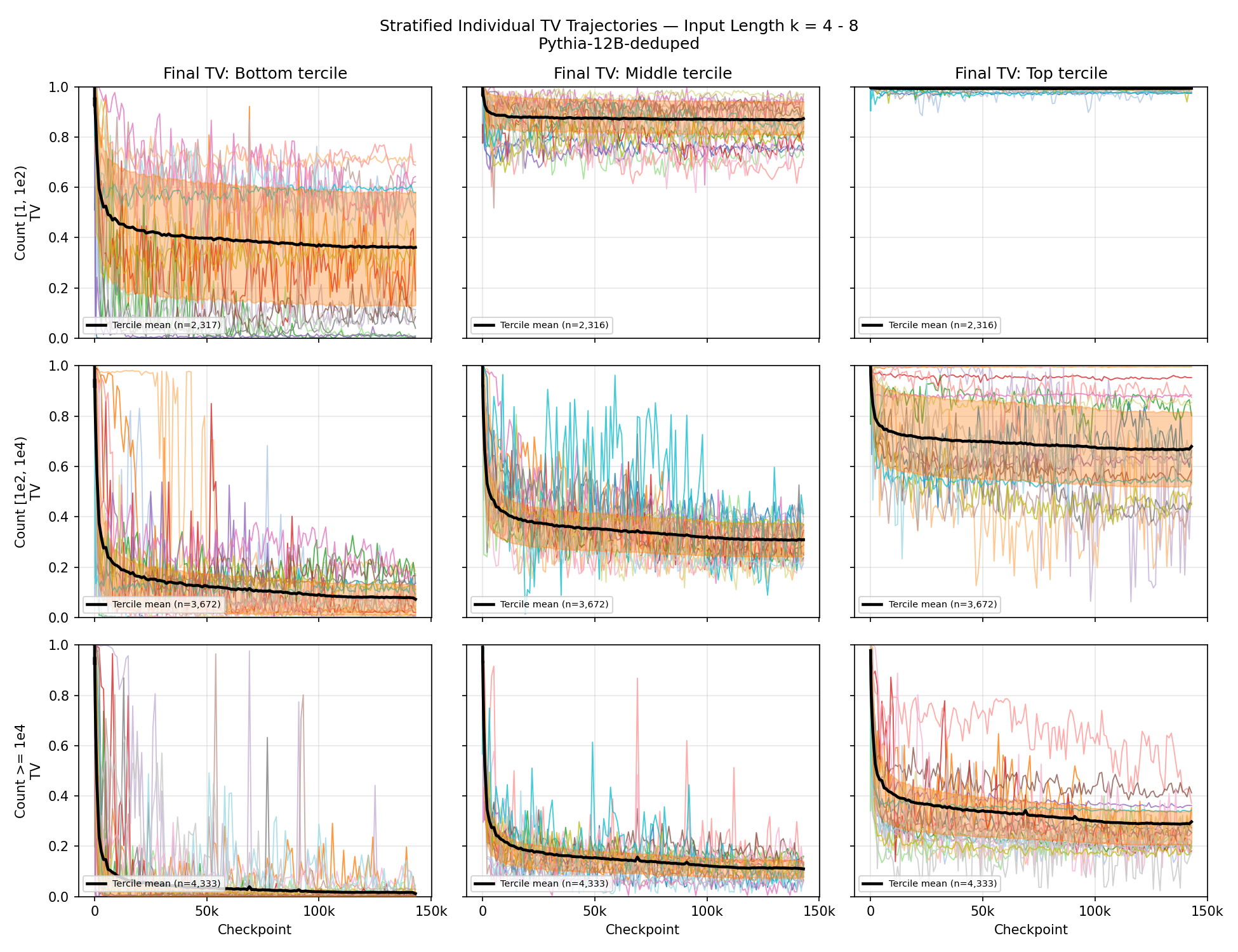}
    \caption{Individual TV trajectories for a small (left) and large (right) model, input length $k=4-8$.}
    \label{fig:individual k4-8}
\end{figure}

\end{document}